 \documentclass[pmlr,twocolumn,10pt]{jmlr} 

\usepackage{booktabs}
\usepackage{siunitx}

\usepackage[switch]{lineno}

\usepackage{multirow}
\usepackage[normalem]{ulem}
\useunder{\uline}{\ul}{}

\theorembodyfont{\upshape}
\theoremheaderfont{\scshape}
\theorempostheader{:}
\theoremsep{\newline}

\jmlrvolume{297}
\jmlryear{2025}
\jmlrworkshop{Machine Learning for Health (ML4H) 2025} 

 \title[Computational Pathology Representational Similarity Analysis]{Comparing Computational Pathology Foundation Models using Representational Similarity Analysis}

\author{%
\Name{Vaibhav Mishra}\Email{vaibhavrichmishra@gmail.com}\\
\addr Dana-Farber Cancer Institute\\
\AND
\Name{William Lotter} \Email{lotterb@ds.dfci.harvard.edu}\\
\addr Dana-Farber Cancer Institute, Brigham and Women's Hospital, \& Harvard Medical School 
}

\begin{document}

\maketitle

\begin{abstract}

Foundation models are increasingly developed in computational pathology (CPath) given their promise in facilitating many downstream tasks. While recent studies have evaluated task performance across models, less is known about the structure and variability of their learned representations. Here, we systematically analyze the representational spaces of six CPath foundation models using techniques popularized in computational neuroscience. The models analyzed span vision-language contrastive learning (CONCH, PLIP, KEEP) and self-distillation (UNI (v2), Virchow (v2), Prov-GigaPath) approaches. Through representational similarity analysis using H\&E image patches from TCGA, we find that UNI2 and Virchow2 have the most distinct representational structures, whereas Prov-Gigapath has the highest average similarity across models. Having the same training paradigm (vision-only vs. vision-language) did not guarantee higher representational similarity. The representations of all models showed a high slide-dependence, but relatively low disease-dependence. Stain normalization decreased slide-dependence for all models by a range of 5.5\% (CONCH) to 20.5\% (PLIP). In terms of intrinsic dimensionality, vision-language models demonstrated relatively compact representations, compared to the more distributed representations of vision-only models. These findings highlight opportunities to improve robustness to slide-specific features, inform model ensembling strategies, and provide insights into how training paradigms shape model representations. Our framework is extendable across medical imaging domains, where probing the internal representations of foundation models can support their effective development and deployment.

\end{abstract}

\paragraph*{Data and Code Availability}
This study leverages the publicly available TCGA dataset. Analysis code is available at \url{https://github.com/lotterlab/cpath_rsa}.

\paragraph*{Institutional Review Board (IRB)}
This research does not require IRB approval. 

\section{Introduction}
\label{sec:intro}

Numerous foundation models have recently been developed for computational pathology. Trained on large-scale datasets in a self-supervised fashion, these models are often designed to generate patch-level embeddings from H\&E-stained whole slide images (WSIs), facilitating downstream tasks such as tumor subtyping, biomarker prediction, and prognostication~\citep{bilal2025review, li2025review, neidlinger2024}. However, recent analyses have also raised questions regarding the robustness of these models to data-specific confounders, such as hospital-specific signatures~\citep{dejong2025, komen2024}. Furthermore, while studies have compared downstream performance across models, the extent to which these models produce similar or divergent embeddings remains unclear. Given the emphasis of interpretability and robustness in medical AI, there is a pressing need to move beyond performance metrics and systematically examine the representations learned by foundation models in computational pathology (CPath).

Here, we leverage techniques from computational neuroscience to explore the representational spaces of six CPath foundation models. Using representational similarity analysis (RSA)~\citep{neuro}, we examine how H\&E-patch embeddings vary across models, diseases, and individual WSIs. The models analyzed represent the two predominant self-supervised learning strategies in the field: vision-language contrastive learning and vision-only self-distillation. As a comparative baseline, we include a DinoV2 model \citep{oquab2024dinov2learningrobustvisual} trained on natural images (i.e., typical, non-medical RGB images such as those in ImageNet). We aimed to quantify similarities and differences between these models, including characterizing the slide- and disease-dependence of their representations, as well as their intrinsic dimensionality. Given the rapidly evolving field, assessing embedding similarity can offers insights into how different training strategies affect the representations learned and guide practical efforts to enhance downstream performance, such as model ensembling and confounder mitigation.

\section{Related Work}

\subsection{Foundation Models in Computational Pathology}

Following strategies developed for natural images, the two most predominant paradigms for developing CPath foundation models are vision-language contrastive learning and vision-only self-distillation~\citep{bilal2025review, li2025review}. Vision-language contrastive learning, pioneered by CLIP~\citep{clip}, aims to learn joint representations between images and text such that matching image-text pairs have similar embeddings and non-matches have dissimilar embeddings. Vision-language models developed for CPath have used various sources to curate image-text pairs, including Twitter (e.g. PLIP~\citep{huang2023visual}), Youtube (e.g., Quilt-1M~\citep{ikezogwo2023quilt1m}), and PubMed (e.g., CONCH~\citep{lu2024avisionlanguage}). KEEP extended traditional vision-language contrastive learning by creating a knowledge graph to enable hierarchical semantic alignment \citep{keep}. For self-distillation, the most commonly adopted algorithm is DinoV2 \citep{oquab2024dinov2learningrobustvisual}, which combines contrastive data augmentation and masking in a student-teacher framework. This learning paradigm aims to learn representations that capture meaningful features of the underlying data that are stable under transformations. CPath models developed using DinoV2 include UNI~\citep{chen2024uni}, Prov-GigaPath~\citep{xu2024gigapath}, and Virchow~\citep{Vorontsov2024, zimmermann2024virchow2}.

\subsection{Analysis of CPath Foundation Models}

Several studies have evaluated the performance of CPath foundation models on downstream tasks such as disease classification, biomarker prediction, and prognostication~\citep{wolflein2024, bilal2025review, neidlinger2024, Bareja2025, marza2025}. In terms of representation analysis, \cite{komen2024} demonstrated that the originating data site can be reliably predicted from the embeddings of several vision-only CPath foundation models, a finding also supported by \cite{dejong2025}.  

\subsection{Representational Similarity Analysis}

Representational similarity analysis (RSA) is a framework used in computational neuroscience to compare neural representations across different conditions, such as comparing the responses of neurons to different stimuli~\citep{neuro}. The core of RSA involves constructing representational dissimilarity matrices (RDMs), which capture pairwise distances between representations across conditions. The RDMs for different systems (e.g., brain regions) can then also be compared to offer insights into similarities and differences between representational structures.

\section{Methods}

\subsection{Models Studied}

We selected six popular foundation models for analysis (Table~\ref{tab:models}), consisting of three vision-only self-distillation models (UNI (v2), Virchow (v2), Prov-Gigapath) and three vision-language models (CONCH, PLIP, KEEP). We additionally include a Dinov2 model trained on natural images as a baseline~\citep{oquab2024dinov2learningrobustvisual}. UNI was trained on over 200 million image patches from over 350k slides sourced from Mass General Brigham~\citep{chen2024uni}. Virchow was trained on 3.1 million WSIs from the Memorial Sloan Kettering Cancer Center~\citep{zimmermann2024virchow2}. For both UNI and Virchow, we use the version 2 models. Prov-GigaPath was trained on 1.3 billion image patches from 171,189 WSIs from the Providence health system~\citep{xu2024gigapath}. CONCH was trained on over 1.17 million image–caption pairs curated from various sources, including PubMed~\citep{lu2024avisionlanguage}. PLIP was trained on 208,414 image–caption pairs curated from public sources (OpenPath), including Twitter~\citep{huang2023visual}. KEEP~\citep{keep} was trained using OpenPath and Quilt-1M~\citep{ikezogwo2023quilt1m}, a dataset of 1M image–caption pairs curated from public sources, including Youtube. All of the studied models use a form of vision transformer (ViT) for the embedder architecture (Table~\ref{tab:models}).

\begin{table*}[h]
\centering 
  \caption{Summary of models studied. }
\begin{tabular}{lccc}
\textbf{Model} & \textbf{Pretraining}              & \textbf{Architecture} & \textbf{Training Data Source} \\
\hline
UNI2            & \multirow{3}{*}{Vision}                 & ViT-H                 & Mass General Brigham          \\
Virchow2        &                                         & ViT-H                 & MSKCC                         \\
Prov-GigaPath  &                                         & ViT-G                 & Providence                    \\
\hline
CONCH          & \multirow{3}{*}{Vision-Language}          & ViT-B                 & Mixed                         \\
PLIP           &                                           & ViT-B                 & OpenPath            \\
KEEP       &                                           & ViT-L                 & Quilt-1M, OpenPath                      \\
\hline
Dinov2       & Vision (Baseline)                            & ViT-B                     & Natural images                     
\end{tabular}
\label{tab:models}
\end{table*}

\subsection{Dataset}

We leverage four subsets of TCGA to perform the analysis: breast invasive carcinoma (BRCA), lung adenocarcinoma (LUAD), lung squamous cell carcinoma (LUSC), and colon adenocarcinoma (COAD)~\citep{tcga}. TCGA was chosen given its common use for benchmarking in computational pathology and because it was not used for pre-training in any of the studied models. Within each cancer subset, 250 whole slide images (WSIs) were randomly sampled to be included in the analysis. For the representational similarity analysis, these WSIs were split into 5 non-overlapping batches of 50 WSIs each. For each WSI, 50 image patches (224x224 pixels) were then sampled following foreground detection using Otsu’s algorithm~\citep{otsu}. Otsu’s method was applied to a grayscale version of the WSI, downsampled by a factor of 224, to identify foreground regions for the patch selection. Altogether the similarity analysis was performed on 50,000 images (4 cancer types $*$ 250 WSIs $*$ 50 patches/WSI), split into 5 batches of 10,000 images each (50 WSIs/batch). The sample size and batching strategy was chosen because of the high computational costs of RSA, which grows according to $N^2$, and to provide a measure of uncertainly as quantified by the range across batches.

\subsection{Embedding Generation}
To evaluate and compare the learned representations across different models, we generated patch-level embeddings for each model for all of the sampled TCGA patches. For each model, we used the specific image-patch processing steps included with the model inference code before passing through the model's encoder to extract the final image representation vector. 

\subsection{Representational Dissimilarity Matrix Generation}
Representational dissimilarity matrix (RDM) construction consists of computing pairwise distances between groups of embeddings. We computed an RDM separately for each model using the Euclidean distance as the distance metric between embedding pairs. The Euclidean distance was used as it is recommended over correlation-based distances when computing RDMs given increased robustness to noise at the representation level \citep{Botero2024, Bosch2025}. When plotting the RDMs, the rows and columns were ordered by WSI and disease type (i.e., all patches from a given WSI are grouped together). RDMs were computed separately for each of the 5 data batches. The rsatoolbox Python package~\citep{10.1371/journal.pcbi.1003553} was used to compute the RDMs. 

\subsection{Comparing RDMs across Models}
To quantify the similarity between model RDMs, we computed pairwise comparisons between all seven models. For each model pair, we loaded the upper-triangular values (excluding the diagonal) of their RDMs and computed the Spearman rank correlation. This process was performed separately for each of the 5 batches, yielding five similarity values per each model pair. 
The results are reported as the mean and range across these values. As an aggregate measure for each model, we also computed the average similarity to all other models. We note that the Spearman correlation is a common metric for comparing RDMs, preferred over Pearson correlation because it does not assume linearity \citep{Bosch2025}. 

A key benefit of RSA is that it abstracts away the differences in embedding dimensionality between models. Specifically, as RSA uses RDMs for model comparison by quantifying pairwise similarities between embeddings across all input images, each RDM has dimensions $N \times N$, where $N$ is the size of the image set, independent of the original embedding size ($d$). These RDMs can then be directly compared across models (e.g. using Spearman correlation) to assess representational similarity. Consequently, RSA enables direct comparison of models with different embedding dimensionalities without requiring dimensional alignment or projection.

To quantitatively group different models according to similarity, we performed hierarchical clustering on the RDM similarity matrices. Ward's method~\citep{Ward01031963} was used to produce a linkage tree, visualized as a dendrogram.

\subsection{Assessing Disease and Slide Specificity}
We assessed the degree to which the embeddings for each model reflect slide-specific or disease-specific information using Cliff’s Delta, a non-parametric statistical test. This analysis was performed on the pairwise distances between the embeddings for different image patches, where, intuitively, if a model's representations are highly slide-dependent, then one would expect that the embeddings from patches from the same slide would be closer than embeddings from different slides. Cliff's Delta was used to provide a measure of this dependence, where 1 indicates a perfect association (i.e., all intra-group embeddings are closer to each other than any inter-group embeddings), and 0 indicates no association.

In more detail, for slide specificity, we computed the distribution of intra-slide distances as pairwise Euclidean distances between all 50 patch embeddings for a given slide, and aggregating across all diseases and slides. Inter-slide distances were computed by measuring Euclidean distances between all combinations of patches between different slides. Cliff's Delta was computed between the intra- and inter-slide distance distributions, and this process was performed separately for each of the 5 batches. For each model, we reported the mean and range of Cliff’s Delta values across batches as an indicator of slide specificity.

To evaluate disease specificity, we computed intra-disease distances by measuring pairwise Euclidean distances among patches of the same subtype, and inter-disease distances between patches of all pairs of different subtypes. This was again repeated across the five batches, and the mean and range of Cliff’s Delta values were reported for each model as an indicator of disease specificity. When computing the set of intra-disease distances, distances between patches from the same WSI were excluded to avoid potential WSI-specific confounding effects.

\subsection{Spectral Analysis}
We quantified the intrinsic dimensionality of the representations for each model using singular value decomposition (SVD). 
We applied SVD to the concatenated embeddings (mean-subtracted) for all slides and disease subtypes for each model. The resulting singular values were normalized to sum to 1, and their cumulative sums were plotted against the percentage of retained features for each model. The rate of increase of these curves as more features are included provides an indicator of the dimensionality of the representational space: a faster increase suggests a relatively low-dimensional structure, whereas a slower increase indicates a more distributed representation in which a larger number of features are required to capture the underlying structure. Given the improved computational efficiency of SVD compare to RSA, we perform SVD across all patches and slides at once for a given model and use 250 samples patches for each slide (250 WSIs $*$ 250 patches/slide $*$ 4 diseases).

\begin{figure*}[h]
\centering 
\includegraphics[width=\textwidth]{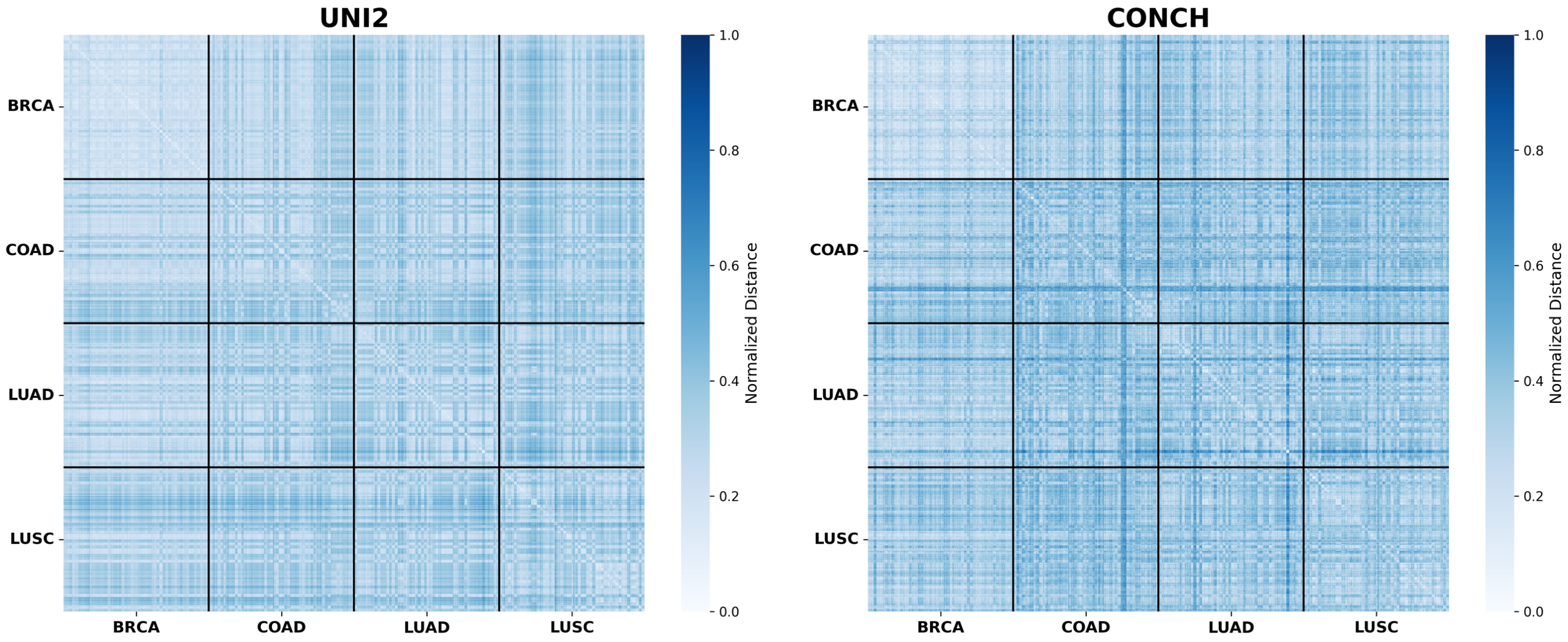} 
\caption{Example Representational Dissimilarity Matrices. The RDMs are computed across 10,000 image patches (50 patches from 50 WSIs for the 4 cancer types) and represent the Euclidean distance between the model representations for each pair of patches, normalized to [0, 1] for each matrix.}
\label{fig:rdm_examples} 
\end{figure*}

\section{Results} 

\subsection{Representational Similarity}

We compared the representations learned between six CPath foundation models using representational similarity analysis, which first involves computing representational dissimilarity matrices (RDMs) for each model separately.
Example RDMs for UNI2 and CONCH are shown in Figure~\ref{fig:rdm_examples}, with RDMs for the remaining models included in the Appendix (Fig.~\ref{fig:all_rdms}).
Each row (column) corresponds to 1 of 10,000 image patches from TCGA, which are ordered by slide (50 patches per slide) and cancer type (50 slides per cancer type).
The values indicate the relative Euclidean distance between the embeddings for each pair of patches.
While $10K * 10K$ = $100M$ pairwise comparisons are inherently complex to visually analyze, some patterns emerge that motivate subsequent quantitative analysis.
For instance, horizontal/vertical streaks appear for groups of adjacent patches (i.e., from the same slide), indicating potential slide-dependent feature representations. 

\begin{figure}[h]
\centering 
\includegraphics[width=\linewidth]{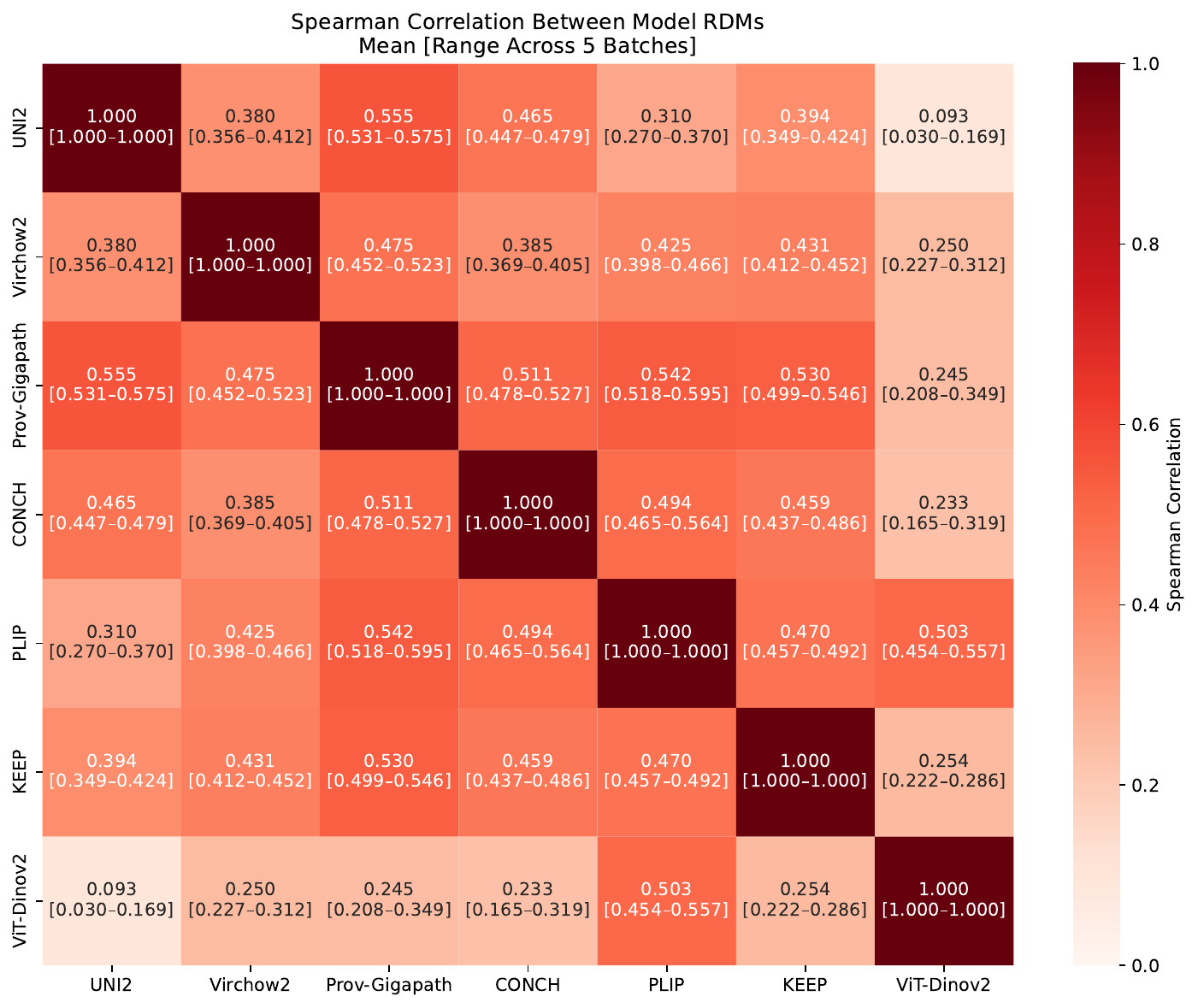} 
\caption{Spearman correlation between the RDMs of each pair of models. The mean and range across the 5 batches are displayed.}
\label{fig:spearman_heatmap} 
\end{figure}

\begin{figure}[h]
\centering 
\includegraphics[width=\linewidth]{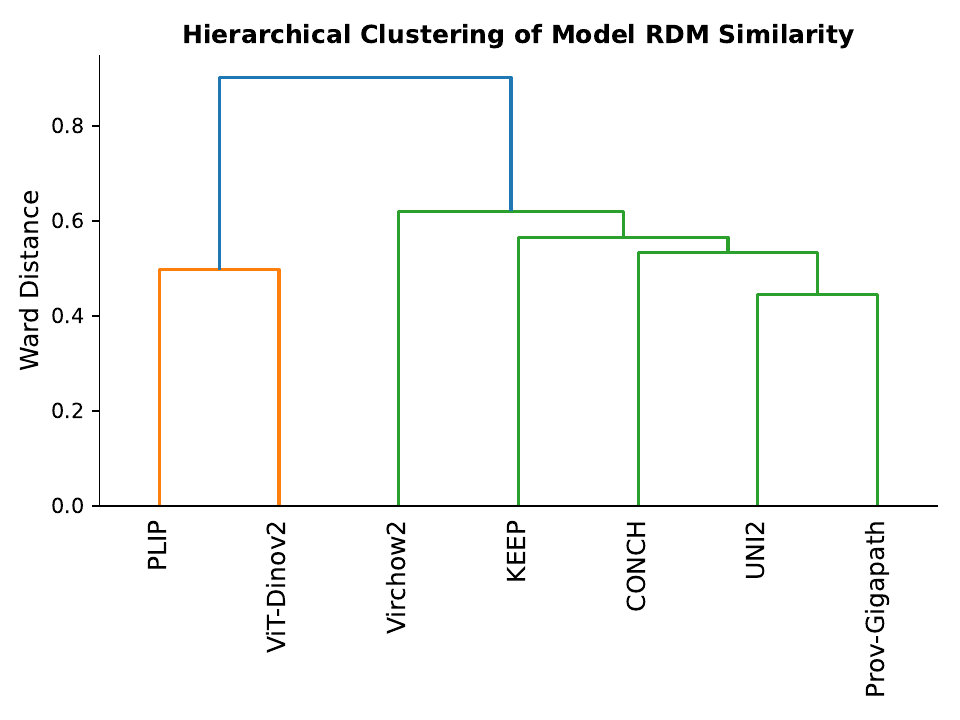} 
\caption{Hierarchical clustering (Ward's method) of the Spearman correlation matrix between the model RDMs.}
\label{fig:spearman_clustering} 
\end{figure}

By converting the representations for each model in a standardized format that is agnostic to feature size, the RDMs enable comparisons of representations across models.
Figure~\ref{fig:spearman_heatmap} summarizes the degree of similarity between each pair of models, quantified as the Spearman correlation between the RDMs of each model.
Higher similarities indicate that image patches with embeddings that are farther apart in one model also tend to be farther apart in the representational space of the other model, indicating common representational structures. 
The highest overall similarity is observed between UNI2 and Prov-Gigapath, with a Spearman correlation of 0.555 between their RDMs (range of 0.531-0.575 across 5 independent batches of $10K$ image patches). 
This finding is intuitive as UNI2 and Prov-Gigapath are both vision-only models trained using the Dinov2 algorithm.
However, not all patterns align with similarities between self-supervised learning strategy.
For instance, out of all of the foundation models, Virchow2's lowest similarity was observed for UNI2, despite both using DinoV2 for training.
On average, Virchow2 and UNI2 were the most distinct in their representational structure, with a mean correlation with the other CPath models of 0.419 and 0.421, respectively (Table~\ref{tab:mean_correlation}).
Prov-GigaPath had the highest average similarity (0.523).
These patterns were consistent across each of the five independent data batches used in this study (each containing 10{,}000 images), where Virchow2 and UNI2 exhibited the lowest average correlations and Prov-Gigapath exhibited the highest correlation within each batch. Paired \textit{t}-tests across batches confirmed that these differences were statistically significant for all comparisons ($p < 0.05$ for all; i.e., Virchow2 $<$ \{Prov-GigaPath, CONCH, PLIP, KEEP\}, UNI2 $<$ \{Prov-GigaPath, CONCH, PLIP, KEEP\}, and Prov-GigaPath $>$ \{UNI2, Virchow2, CONCH, PLIP, KEEP\}).
The baseline ViT-Dinov2 trained on natural images had low similarity ($\le$0.3) to each model except PLIP (0.503). Hierarchical clustering on the Spearman similarity matrix revealed similar patterns (Figure~\ref{fig:spearman_clustering}), with ViT-Dinov2 and PLIP separating from the other foundation models. 

\begin{table}[h]
\centering 
\small
  \caption{Mean Spearman correlation between the RDM for each model and the other five CPath models. Bold and underline represent the highest and lowest values, respectively.}
\begin{tabular}{lcc}
\textbf{Model} & \textbf{Pretraining}             & \textbf{Mean Corr.} \\
\hline
UNI2            & \multirow{3}{*}{Vision}          & 0.421                     \\
Virchow2        &                                  & {\ul 0.419}                     \\
Prov-Gigapath  &                                  & \textbf{0.523}                     \\
\hline
CONCH          & \multirow{3}{*}{Vision-Language} & 0.463                     \\
PLIP           &                                  & 0.448                     \\
KEEP       &                                  & 0.457                     \\
\hline
\end{tabular}
\label{tab:mean_correlation}
\end{table}

\subsection{Disease- and Slide- Specificity}
In addition to model similarity, we evaluated the degree of slide- and disease-specificity for each model by comparing the relative distances between embeddings from the same slide and different slides, and separately from the same disease and different diseases.
Cliff's Delta, a non-parametric measure of the magnitude of difference between two groups, was used to quantify these effects.
Ranging from 0 to 1, a Cliff's Delta greater than 0.43 is regarded as a large effect, with values of 0.28 and 0.11 indicating medium and weak effects, respectively~\citep{vargha}.

\begin{table*}[t]
\centering 
  \caption{Assessment of slide- and disease-specificity of model representations. Cliff's Delta, a non-parametric test of effect size that ranges between 0 and 1, was computed based on the RDMs for each model. The mean and range across the 5 batches is displayed, where higher values indicate higher specificity. Bold and underline represent the highest and lowest values for the CPath models, respectively.}
\begin{tabular}{lcccc}
\textbf{}      & \multicolumn{2}{c}{\textbf{Slide Specificity}} & \multicolumn{2}{c}{\textbf{Disease Specificity}} \\
\textbf{Model} & \textbf{Cliff’s Delta}   & \textbf{Range}      & \textbf{Cliff’s Delta}    & \textbf{Range}       \\
\hline
UNI2            & 0.751         & {[}0.685, 0.780{]}  & 0.144                     & {[}0.076, 0.182{]}   \\
Virchow2        & 0.615                    & {[}0.551, 0.667{]}  & {\ul 0.120}                & {[}0.058, 0.162{]}   \\
Prov-Gigapath  & \textbf{0.762}                    & {[}0.658, 0.807{]}  & 0.138                     & {[}0.066, 0.169{]}   \\
\hline
CONCH          & {\ul 0.596}              & {[}0.501, 0.636{]}  & 0.135                     & {[}0.072, 0.166{]}   \\
PLIP           & 0.700                    & {[}0.601, 0.748{]}  & 0.152                     & {[}0.073, 0.192{]}   \\
KEEP       & 0.703                    & {[}0.618, 0.732{]}  & \textbf{0.210}            & {[}0.138, 0.242{]}   \\
\hline
ViT-Dinov2       & 0.367                    & {[}0.315, 0.411{]}  & 0.066                     & {[}0.029, 0.091{]}  
\end{tabular}
\label{tab:specificity}
\end{table*}

The results of the slide- and disease-specificity analyses are contained in Table~\ref{tab:specificity}.
All of the CPath models demonstrate relatively large slide-specificity, indicating that embeddings from patches from the same WSI tend to be closer together than embeddings from different WSIs.
The highest level of slide specificity was observed for Prov-Gigapath, with a Cliff's Delta of 0.762 (range: 0.658-0.807 across the 5 batches), followed closely by UNI2 (0.751, [0.685, 0.780]).
CONCH demonstrated the lowest value of 0.596 [0.501, 0.636].
All of the CPath models had a higher slide-specificity than the ViT-Dinov2 baseline (0.367).
For disease-specificity, the associations were weak for each of the models, with a range of 0.120 (Virchow) to 0.210 (KEEP). 
Thus, while embeddings from the same disease tended to be slightly closer together than embeddings from different diseases, the representations had a much stronger dependency on individual slides.
No clear trends were observed regarding vision-language versus vision-only models for either slide- or disease-specificity.

\subsection{Spectral Analysis}
\begin{figure}[h]
\centering 
\includegraphics[width=\linewidth]{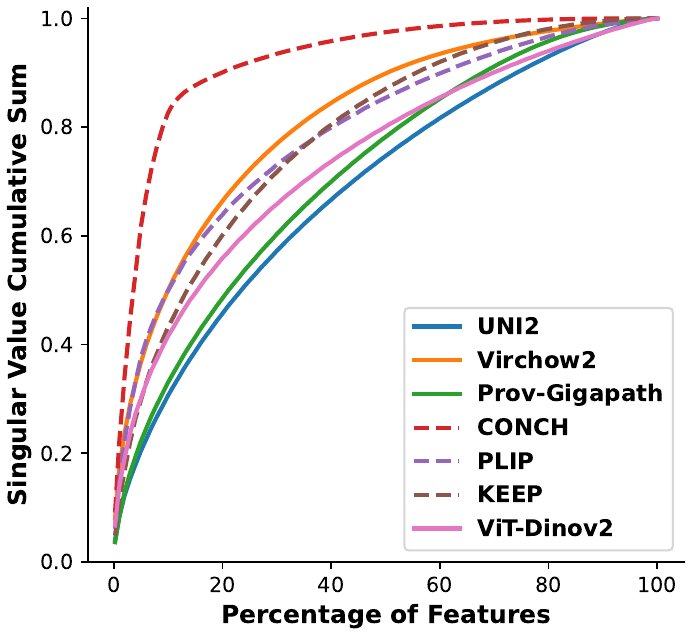} 
\caption{Spectral analysis. Singular value decomposition was performed on the foundation model representations. Singular values were normalized to sum to 1 and the cumulative sum is plotted with respect to the percentage of features included. Solid lines indicate vision-only models; dashed lines indicate vision-language models.}
\label{fig:spectral_analysis} 
\end{figure}

Finally, we performed spectral analysis to investigate the intrinsic dimensionality of the feature representations for each model.
Figure~\ref{fig:spectral_analysis} illustrates the cumulative distribution of singular values as a function of the percentage of features for each model. As indicated by sharper rises in the curves, the vision-language models appear to have relatively lower dimensional representations, especially CONCH.
Conversely, the curves of UNI2 and Prov-Gigapath have the most shallow rises, indicating relatively distributed representations with higher dimensionality. 

\subsection{Effects of Stain Normalization}
Given the observed high slide-specificity, we examined whether stain normalization -- absent from the default processing of all models -- would influence the results. Each patch was normalized using Macenko normalization \citep{macenko} using the TiaToolbox \citep{tiatoolbox}, and the analysis steps were repeated using the normalized patches. Normalization had little effect on the RDM correlation structure between models (Figure \ref{fig:spearman_heatmap-normalized} and Table \ref{tab:mean_correlation_normalized} in Appendix). However, the slide-specificity of the representations decreased for each model (Table \ref{tab:specificity_normalized} in Appendix). The largest effect was observed for PLIP (20.5\%), with CONCH showing the smallest effect (5.5\% decrease). Virchow2 showed the lowest slide-specificity after normalization, with Prov-Gigapath remaining the highest. The disease-specificity of the representations also decreased with normalization, though to a lesser degree in magnitude.

\subsection{Hyperparameter Sensitivity}
To assess robustness to the patch sampling configuration, we repeated all analyses using a different number of patches per slide. The original results were computed over five independent batches of 50 WSIs × 50 patches/WSI (× 4 diseases = 10,000 images). The new sampling configuration used five independent batches of 25 WSIs × 100 patches/WSI (× 4 diseases = 10,000 images).
We observed highly similar results: the Pearson correlation was 0.997 between the original model-similarity values (Figure \ref{fig:spearman_heatmap}) and the new values (Figure \ref{fig:100/25-spearman_heatmap}), with UNI2 and Virchow2 again exhibiting the lowest average correlations and Prov-GigaPath the highest. The Pearson correlations for slide-specificity and disease-specificity were 0.999 and 0.984, respectively (Table~\ref{tab:specificity}, Table~\ref{tab:specificity_25_100}).

The original analyses used Euclidean distance for RDM construction, recommended for its robustness to representational noise~\citep{Botero2024,Bosch2025}. Because several models were trained with contrastive loss, we additionally evaluated whether using a correlation-based metric would alter the RSA results. Using Pearson correlation for RDM construction yielded similar findings: Virchow2 again exhibited the lowest average similarity across models and Prov-GigaPath the highest. The Pearson correlation between the full set of original similarity values between models (Figure \ref{fig:spearman_heatmap}) and the new values (Figure \ref{fig:Pearson-correlation-spearman_heatmap}) was 0.84, indicating high correlation with some expected quantitative variation. In particular, UNI2’s average similarity between models increased from 0.421 to 0.527. Thus, while there are some quantitative changes with a different distance metric, the overall findings remain robust. We note that the additional experiments (slide/disease specificity, spectral analysis) are independent of the RDM construction. 

\section{Discussion} 
Beyond assessing performance in downstream tasks, it is important to investigate the representations learned by foundation models, which can offer insights into the strengths and weaknesses of current approaches and the influence of training strategies on embedding structure. 
Using techniques popularized in computational neuroscience, we compared the representations of six CPath models and a baseline natural image model across 50,000 image patches and four cancer types.

We find that UNI2 and Virchow2 are the most distinct in their representations, as quantified by the average similarity between model RDMs. Having the same pre-training scheme did not guarantee higher similarity between models, as the model most similar to CONCH and KEEP (both vision-language) was Prov-GigaPath (vision-only) and the model most dissimilar to Virchow2  was UNI2 (both vision-only).
The causes underlying these findings are unclear, but could potentially relate to differences in datasets or hyperparameters. Future work could systematically analyze how variations in these parameters affect representational structure.

One notable observation is the higher correlation between PLIP and the natural image baseline (ViT-DINOv2). One possible explanation is that PLIP was trained on fewer pathology images (208K) than the other models (each over 1M). As a result, its representations may partially overlap with the broader visual embeddings of natural images. Nevertheless, this does not preclude strong performance in certain tasks, such as PLIP performing well on zero-shot classification in external datasets~\citep{huang2023visual}, highlighting that representational overlap with natural images can still support robust transfer in clinically relevant tasks.

Using the distances between representations, we also quantified the degree of slide- and disease-specificity of each model.
All of the CPath models demonstrated a high slide-specificity, where embeddings from patches from the same slide tended to be closer together than embeddings from different slides.
This is partly expected and could be beneficial for certain tasks -- some morphological features would be expected to vary more across tissue samples than within tissue samples, and capturing these variations could be important in some precision medicine tasks. However, the reduction in slide-specificity caused by stain normalization also suggests some lack of robustness to slide-specific confounders. 
Conversely, the models demonstrated relatively low disease specificity.
As foundation models have generally shown high downstream performance in tumor type classification~\citep{Wang2024-qz, Bareja2025}, this result is potentially surprising. 
However, these findings are not necessarily at odds, where certain combinations of features may remain stable across patches/WSIs for a given tumor type, even if there are variations in the representations as a whole.

In terms of intrinsic dimensionality, we found that the vision-language models had relatively lower-dimensional representations compared to the relatively distributed representations of vision-only models. One potential explanation relates to differences in training objectives between these approaches. Vision-language models are trained to align image embeddings with corresponding language embeddings, introducing a language bottleneck wherein visual representations are forced to align with lower-dimensional, semantically structured language embeddings in a shared multimodal space. This alignment may thus encourage compression of rich visual details into more compact representations. In contrast, current vision-only models use image-based objectives based on masking and data augmentation, which may promote the preservation of richer visual details. 

Another interesting observation was that Prov-GigaPath exhibited high representational similarity across models, high slide specificity, and high intrinsic dimensionality These observations are not mutually exclusive and may, in fact, be related. For example, because all models demonstrate high slide specificity, a model with the greatest slide specificity may then show higher representational similarity to other models. Additionally, higher intrinsic dimensionality may correlate with capturing more fine-grained, slide-specific features. Consistent with this interpretation, the lowest intrinsic dimensionality was observed for CONCH, which also exhibited the lowest slide specificity, followed by Virchow2, which had the second lowest values for both.

Our findings have several implications for the development and deployment of foundation models in computational pathology. The observed high slide-specificity and effects of stain normalization suggests some lack of robustness in current models, consistent with other recent studies~\citep{komen2024, dejong2025}.
Using data augmentation and/or adversarial learning during model development could potentially improve this robustness \citep{dann, advdino}, and stain normalization may help at inference time.
The existence of slide-specific representational signatures is not necessarily detrimental to downstream tasks, but it does raise the possibility of shortcut learning and reduced generalization when fitting classifiers on top of the embeddings.
Additionally, our findings regarding the pairwise similarities between models can inform ensembling strategies. 
Ensembling of different models often improves downstream performance, yet the high computational costs of CPath foundation models limits the number of models that can be ensembled.
Thus, ensembling could focus on more dissimilar, complementary models.

Several recent works have provided extensive downstream evaluation of pathology foundation models \citep{marza2025, Bareja2025}. One intriguing potential connection between those results and our findings relates to Virchow2. \cite{Bareja2025} found that vision-only models outperformed other model types and Virchow2 had the highest overall performance of all models. In our study, we also found that Virchow2 had lower slide specificity than the other vision-only models. While this is not a causal connection, it suggests potential links to motivate follow-up work, where our study provides a unique and complementary perspective compared to other efforts in the field.

\paragraph{Limitations}
While we studied four disease cohorts, all of these cohorts came from TCGA, so the results may quantitatively differ in different datasets. More broadly, it will be valuable in future work to explicitly assess generalizability across different institutions, staining protocols, and scanner hardware, including how the studied representational metrics relate to downstream performance. 
Another challenge of RSA is its high CPU memory demand, which scales quadratically with the number of samples ($N^2$). This places constraints on sample size and can require high-memory computing nodes. 

\paragraph{Conclusion}
As foundation models are increasingly developed and applied across medical domains, the analytical approach presented here may be effectively extended to these areas. 
Gaining insight into the structure of learned representations can ultimately inform algorithmic refinement and support safe, effective deployment in clinical settings.

\acks{W.L. acknowledges funding support from the Ellison Foundation, the Wong Family Award, the Louis B. Mayer Foundation, the National Institute of Biomedical Imaging and Bioengineering award R21EB035247, and the National Library of Medicine award R01LM014775.}

\bibliography{main.bib}

\clearpage
\appendix

\section*{Appendix}\label{apd:first}

\begin{figure}[h]
\centering 
\includegraphics[width=\linewidth]{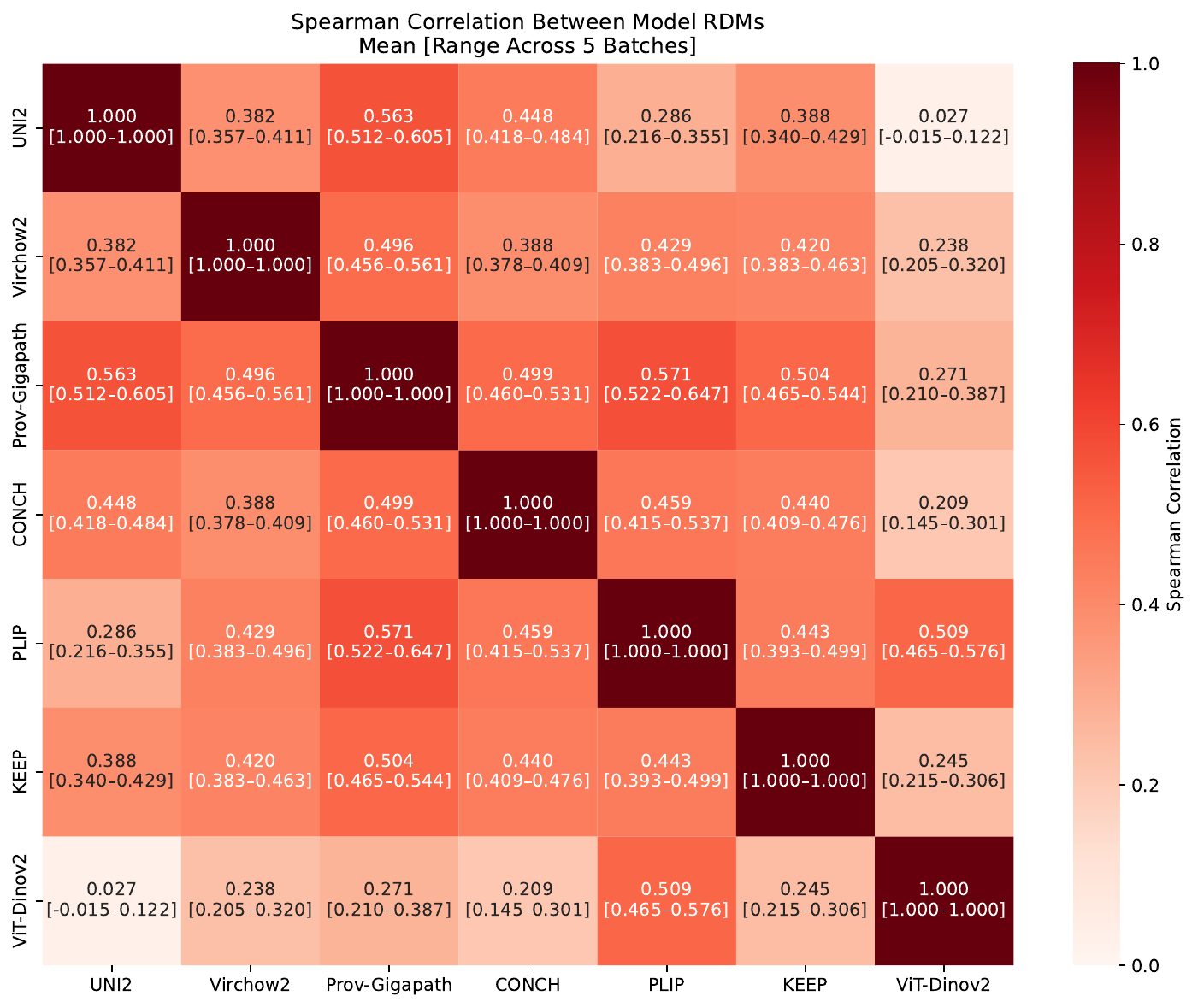} 
\caption{Spearman correlation between the RDMs of each pair of models using stain-normalized patches. Analogous to Figure \ref{fig:spearman_heatmap} except using the stain-normalized patches.}
\label{fig:spearman_heatmap-normalized} 
\end{figure}

\begin{figure}[h]
\centering 
\includegraphics[width=\linewidth]{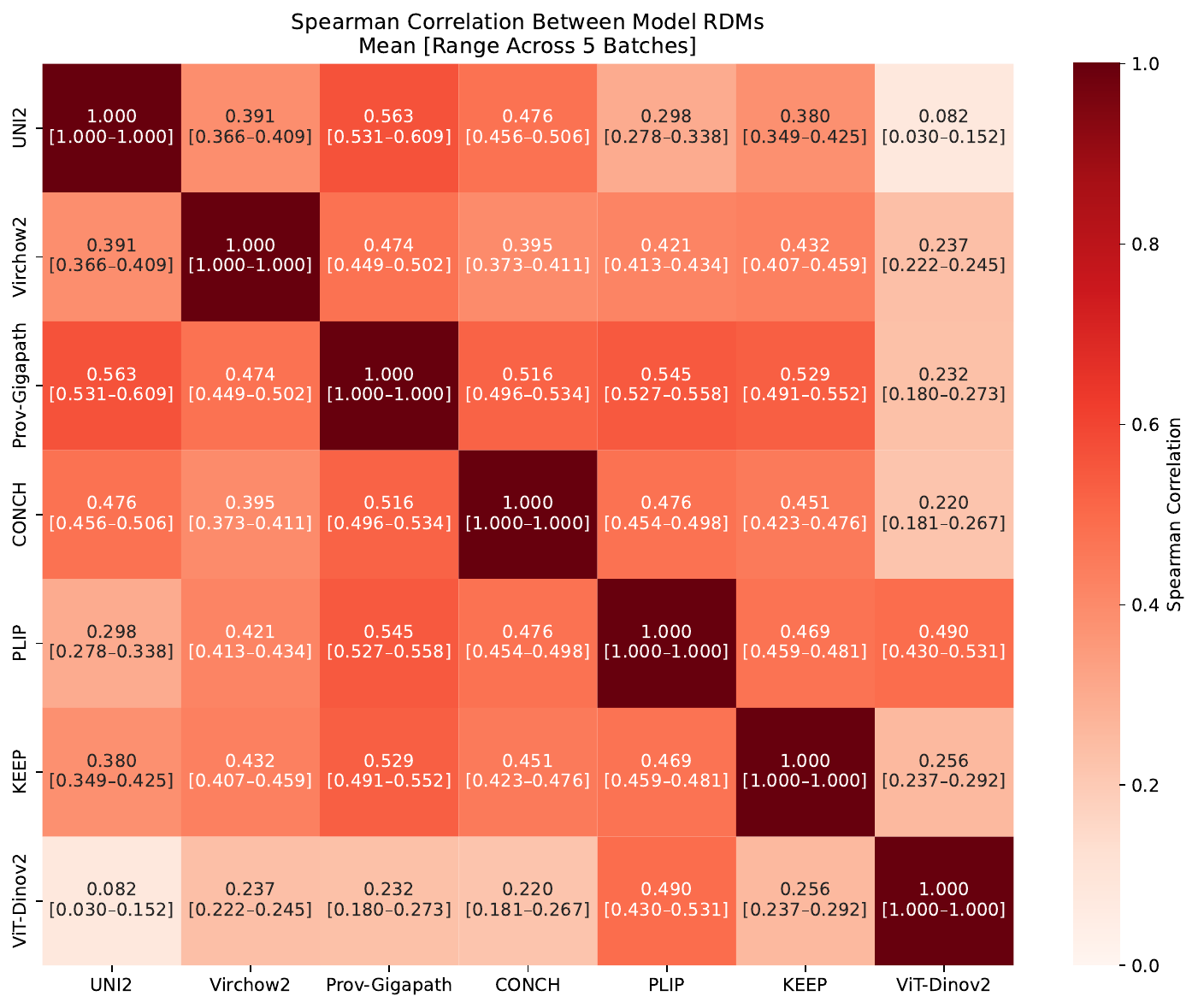} 
\caption{Spearman correlation between the RDMs of each pair of models using 25 WSIs per disease with 100 patches per WSI. Analogous to Figure \ref{fig:spearman_heatmap} except using the modified sampling scheme.}
\label{fig:100/25-spearman_heatmap} 
\end{figure}

\begin{figure}[h]
\centering 
\includegraphics[width=\linewidth]{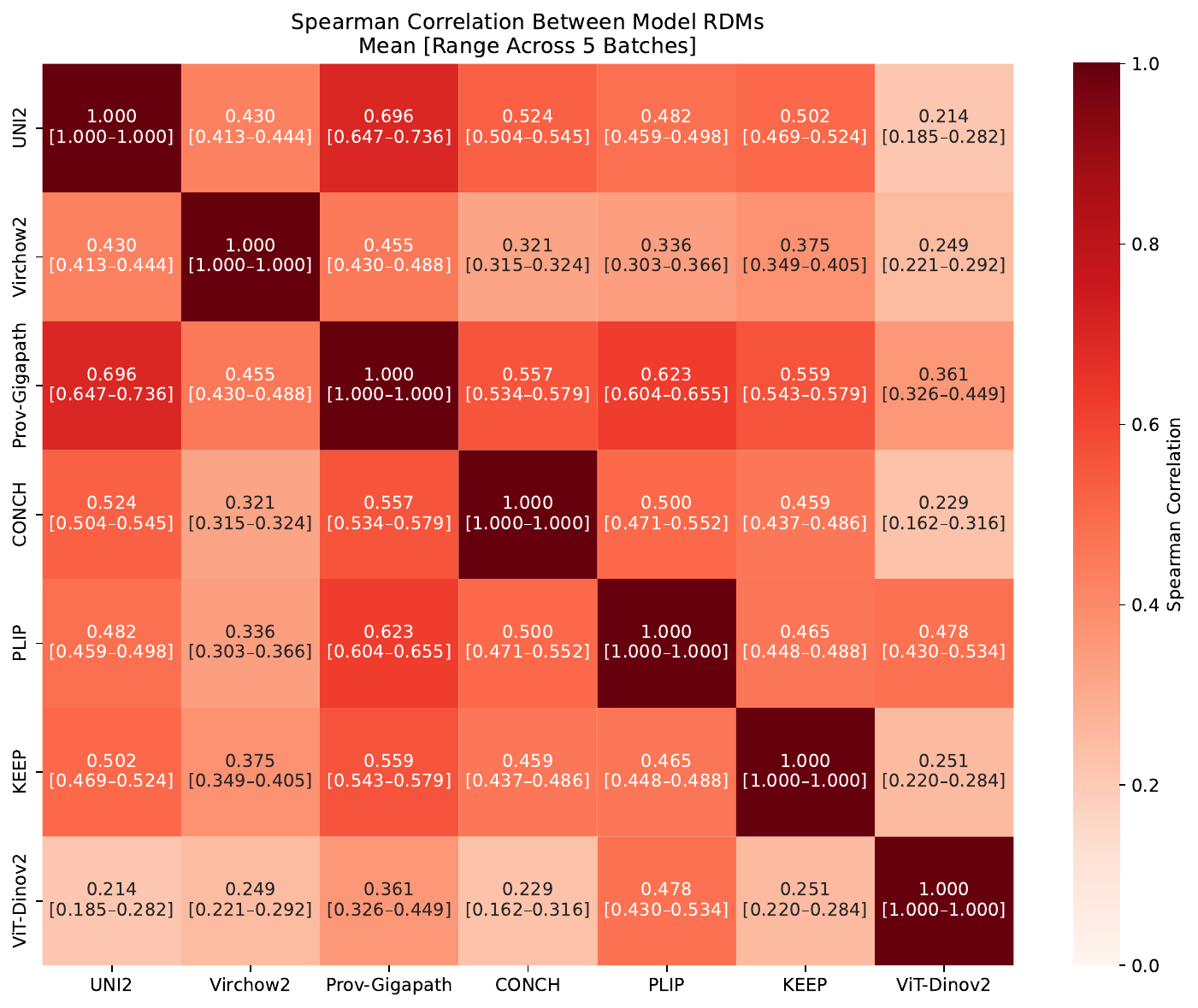} 
\caption{Spearman correlation between the RDMs of each pair of models using Pearson correlation distance for RDM construction instead of Euclidean. Analogous to Figure \ref{fig:spearman_heatmap} except using the different distance metric.}
\label{fig:Pearson-correlation-spearman_heatmap} 
\end{figure}

\begin{table}[h]
\centering 
\small
  \caption{Mean Spearman correlation between the RDM for each model and the other five CPath models using the stain-normalized patches. Analogous to Table \ref{tab:mean_correlation} except using the stain-normalized patches.}
\begin{tabular}{lcc}
\textbf{Model} & \textbf{Pretraining}             & \textbf{Mean Corr.} \\
\hline
UNI2            & \multirow{3}{*}{Vision}          & {\ul 0.413}                     \\
Virchow2        &                                  & 0.423                     \\
Prov-Gigapath  &                                  & \textbf{0.527}                     \\
\hline
CONCH          & \multirow{3}{*}{Vision-Language} & 0.447                     \\
PLIP           &                                  & 0.438                     \\
KEEP       &                                  & 0.439                     \\
\hline
\end{tabular}
\label{tab:mean_correlation_normalized}
\end{table}

\begin{table}[h]
\centering 
\small
  \caption{Mean Spearman correlation between the RDM for each model and the other five CPath models using 25 WSIs per disease with 100 patches per WSI. Analogous to Table \ref{tab:mean_correlation} except using this modified sampling scheme.}
\begin{tabular}{lcc}
\textbf{Model} & \textbf{Pretraining}             & \textbf{Mean Corr.} \\
\hline
UNI2            & \multirow{3}{*}{Vision}          & {\ul 0.422}                     \\
Virchow2        &                                  & 0.423                     \\
Prov-Gigapath   &                                  & \textbf{0.525}                     \\
\hline
CONCH           & \multirow{3}{*}{Vision-Language} & 0.463                     \\
PLIP            &                                  & 0.442                     \\
KEEP            &                                  & 0.452                     \\
\hline
\end{tabular}
\label{tab:mean_correlation_100/25}
\end{table}

\begin{table}[h]
\centering 
\small
  \caption{Mean Spearman correlation distance between the RDM for each model and the other five CPath models when using Pearson correlation distance for RDM construction. Analogous to Table \ref{tab:mean_correlation} except using the different distance metric.}
\begin{tabular}{lcc}
\textbf{Model} & \textbf{Pretraining}             & \textbf{Mean Corr.} \\
\hline
UNI2            & \multirow{3}{*}{Vision}          & 0.527                     \\
Virchow2        &                                  & {\ul 0.383}                     \\
Prov-Gigapath   &                                  & \textbf{0.578}                     \\
\hline
CONCH           & \multirow{3}{*}{Vision-Language} & 0.472                     \\
PLIP            &                                  & 0.481                     \\
KEEP            &                                  & 0.472                     \\
\hline
\end{tabular}
\label{tab:pearson_correlation_distance}
\end{table}

\begin{table*}[h]
\centering 
  \caption{Assessment of slide- and disease-specificity of model representations using stain-normalized patches. Analogous to Table \ref{tab:specificity} except using the stain-normalized patches. $\Delta$ Orig. represents the change compared to the original, unnormalized patches.}
\begin{tabular}{lcccccc}
\textbf{}      & \multicolumn{3}{c}{\textbf{Slide Specificity}} & \multicolumn{3}{c}{\textbf{Disease Specificity}} \\
\textbf{Model} & \textbf{Cliff’s Delta}   & \textbf{Range} & \textbf{$\Delta$ Orig.}     & \textbf{Cliff’s Delta}    & \textbf{Range}   & \textbf{$\Delta$ Orig.}    \\
\hline
UNI2            & 0.692         & {[}0.631, 0.724{]} & -0.059 & 0.112                     & {[}0.066, 0.143{]} & -0.033  \\
Virchow2        & {\ul 0.547}                    & {[}0.473, 0.585{]} & -0.068 & {\ul 0.084}                & {[}0.039, 0.104{]}  & -0.036 \\
Prov-Gigapath  & \textbf{0.712}                    & {[}0.597, 0.761{]} & -0.050 & 0.124                     & {[}0.051, 0.165{]} & -0.014  \\
\hline
CONCH          & 0.563              & {[}0.477, 0.610{]} & -0.033 & 0.112                     & {[}0.055, 0.138{]}  & -0.023 \\
PLIP           & 0.557                    & {[}0.471, 0.590{]} & -0.143 & 0.093                     & {[}0.038, 0.112{]}  & -0.059 \\
KEEP       & 0.613                    & {[}0.526, 0.651{]} & -0.090 & \textbf{0.169}            & {[}0.107, 0.200{]}  & -0.041 \\
\hline
ViT-Dinov2       & 0.326                    & {[}0.286, 0.349{]} & -0.042 & 0.059                     & {[}0.027, 0.080{]}  & -0.006
\end{tabular}
\label{tab:specificity_normalized}
\end{table*}

\begin{table*}[t]
\centering 
  \caption{Assessment of slide- and disease-specificity of model representations when sampling 25 WSIs per disease with 100 patches per WSI. Analogous to Table \ref{tab:specificity} except using the different sampling parameters.}
\begin{tabular}{lcccc}
\textbf{}      & \multicolumn{2}{c}{\textbf{Slide Specificity}} & \multicolumn{2}{c}{\textbf{Disease Specificity}} \\
\textbf{Model} & \textbf{Cliff’s Delta}   & \textbf{Range}      & \textbf{Cliff’s Delta}    & \textbf{Range}       \\
\hline
UNI2            & 0.773         & {[}0.742, 0.804{]}  & 0.196                     & {[}0.163, 0.216{]}   \\
Virchow2        & 0.639                    & {[}0.582, 0.673{]}  & {\ul 0.163}                & {[}0.140, 0.192{]}   \\
Prov-Gigapath   & \textbf{0.795}                    & {[}0.778, 0.808{]}  & 0.184                     & {[}0.159, 0.205{]}   \\
\hline
CONCH           & {\ul 0.629}              & {[}0.612, 0.645{]}  & 0.174                     & {[}0.157, 0.187{]}   \\
PLIP            & 0.730                    & {[}0.710, 0.745{]}  & 0.190                     & {[}0.129, 0.236{]}   \\
KEEP            & 0.728                    & {[}0.698, 0.750{]}  & \textbf{0.245}            & {[}0.240, 0.252{]}   \\
\hline
ViT-Dinov2      & 0.388                    & {[}0.336, 0.415{]}  & 0.088                     & {[}0.070, 0.099{]}  
\end{tabular}
\label{tab:specificity_25_100}
\end{table*}

\begin{figure*}[h]
\centering 
\includegraphics[width=0.75\textwidth]{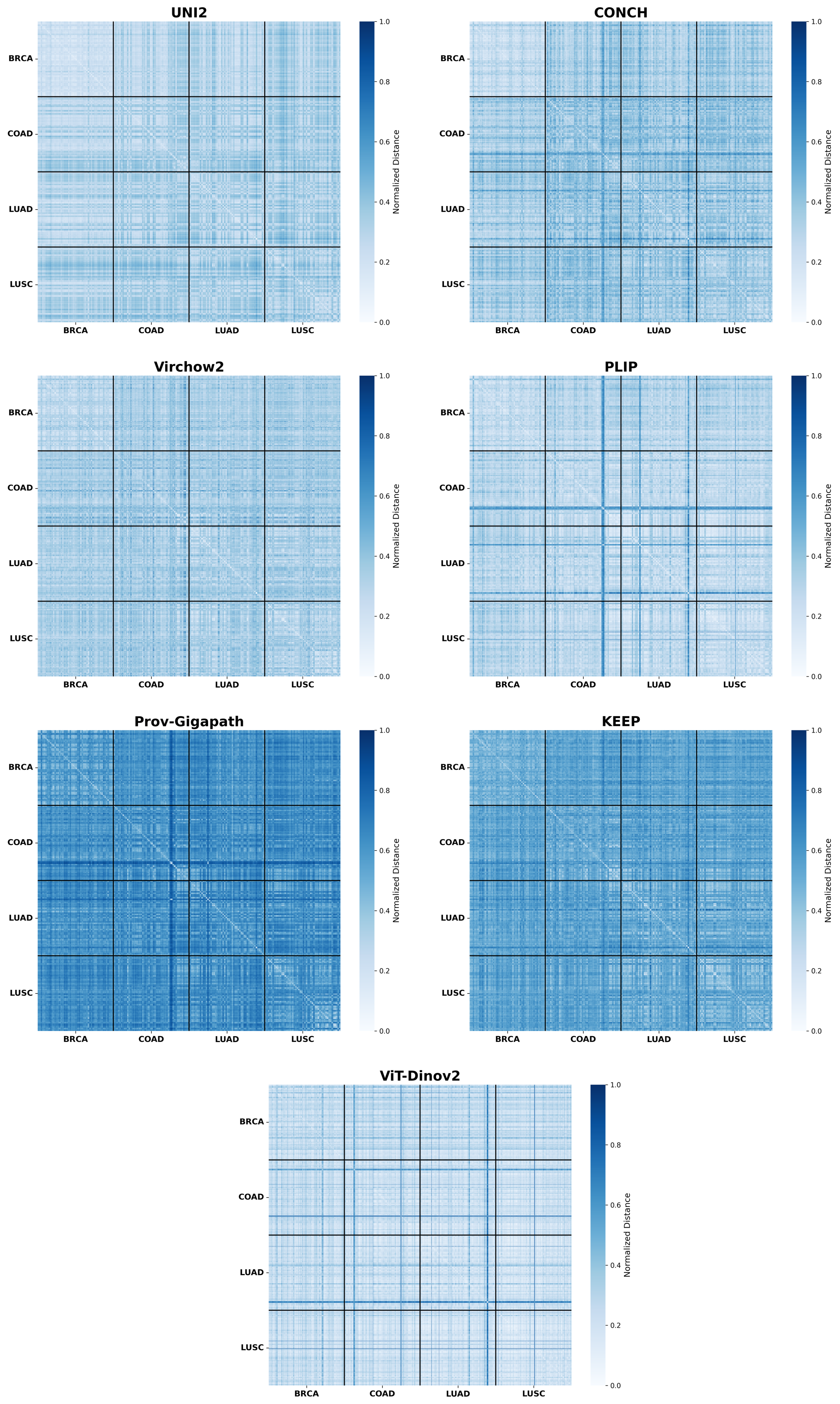} 
\vspace{-10pt}
\caption{Representational Dissimilarity Matrices (RDMs) for all models.}
\label{fig:all_rdms} 
\end{figure*}



\end{document}